\documentclass[sigconf]{acmart}
\AtBeginDocument{%
  \providecommand\BibTeX{{%
    Bib\TeX}}}
\usepackage{listings}
\usepackage{circledtext}
\usepackage{tikz}
\usepackage{caption}
\usepackage{subcaption}
\usepackage{graphicx}  
\usepackage{booktabs}
\usepackage{fancyhdr}

\copyrightyear{2026}
\acmYear{2026}
\setcopyright{cc}
\setcctype{by}
\acmConference[WWW '26]{Proceedings of the ACM Web Conference 2026}{April 13--17, 2026}{Dubai, United Arab Emirates}
\acmBooktitle{Proceedings of the ACM Web Conference 2026 (WWW '26), April 13--17, 2026, Dubai, United Arab Emirates}
\acmPrice{}
\acmDOI{10.1145/3774904.3792864}
\acmISBN{979-8-4007-2307-0/2026/04}

\newenvironment{compactItemizeEmptyBullet}{\begin{list}{$\circ$}
			{\setlength{\topsep}{1mm}\setlength{\itemsep}{0.25mm}
				\setlength{\parsep}{0.1mm}
				\setlength{\itemindent}{0mm}\setlength{\partopsep}{0mm}
				\setlength{\labelwidth}{15mm}
				\setlength{\leftmargin}{4mm}}}{\end{list}}

\def\BibTeX{{\rm B\kern-.05em{\sc i\kern-.025em b}\kern-.08em
    T\kern-.1667em\lower.7ex\hbox{E}\kern-.125emX}}
\newenvironment{compactItemize}{\begin{list}{$\bullet$}
			{\setlength{\topsep}{1mm}\setlength{\itemsep}{0.25mm}
				\setlength{\parsep}{0.1mm}
				\setlength{\itemindent}{0mm}\setlength{\partopsep}{0mm}
				\setlength{\labelwidth}{15mm}
				\setlength{\leftmargin}{4mm}}}{\end{list}}

\newcounter{myctr}

\begin{document}

\fancypagestyle{firststyle}
{
   \fancyhf{}
   \fancyhead[C]{This is an authors' copy of the paper  to appear in  Companion Proceedings of the ACM Web Conference
2026 (WWW Companion '26).}
   \renewcommand{\headrulewidth}{0pt} 
}

\title{Access Controlled Website Interaction for Agentic AI with Delegated Critical Tasks}


\author{Sunyoung Kim}
\affiliation{%
  \institution{Arizona State University}
  \city{Tempe}
  \state{AZ}
  \country{USA}}
\email{skim638@asu.edu}
\orcid{0009-0002-7059-5911}

\author{Hokeun Kim}
\affiliation{%
  \institution{Arizona State University}
  \city{Tempe}
  \state{AZ}
  \country{USA}}
\email{hokeun@asu.edu}
\orcid{0000-0003-1450-5248}

\renewcommand{\shortauthors}{Sunyoung Kim and Hokeun Kim}

\begin{abstract}
Recent studies reveal gaps in delegating critical tasks to agentic AI that accesses websites on the user's behalf, primarily due to limited access control mechanisms on websites designed for agentic AI. In response, we propose a design of website-based interaction for AI agents with fine-grained access control for delegated critical tasks. Our approach encompasses a website design and implementation, as well as modifications to the access grant protocols in an open-source authorization service to tailor it to agentic AI, with delegated critical tasks on the website. The evaluation of our approach demonstrates the capabilities of our access-controlled website used by AI agents.
\vspace{-5pt}
\end{abstract}

\ccsdesc[500]{Security and privacy~Access control}
\ccsdesc[500]{Information systems~Web applications}
\ccsdesc[500]{Computing methodologies~Natural language generation}

\keywords{Access control, Delegation, Website design, Agentic AI}


\maketitle
\thispagestyle{firststyle}
\vspace{-5pt}
\section{Introduction}
Recent advances in agentic artificial intelligence (agentic AI) have enabled autonomous systems to perform complex, critical tasks such as
business operations and e-commerce~\cite{pymnts2024}, algorithmic trading~\cite{zhang2024multimodal}, or healthcare~\cite{moritz2025coordinated}.
The release of OpenAI's Agent Kit~\cite{openai2025} has brought agentic AI to everyday life, making it more accessible for both developers and regular users.
These AI agents act on behalf of users, accessing web services and performing critical tasks. 


However, the absence of robust access control of agentic AI presents severe risks, such as unauthorized purchases, identity theft, and data exposure~\cite{khan2024security}.
Thus, research on agentic AI has increasingly focused on access control.
A recent study~\cite{huang2025novel} proposes zero-trust models to manage agent identities in multi-agent systems, and extend authentication protocols such as OpenID Connect to enable verified delegation~\cite{south2025authenticated}. 
Major financial providers have also begun exploring this area. 
For example, Mastercard’s Agent Pay~\cite{mastercard2025} enables AI agents to make real-world payments on behalf of users using tokenization~\cite{mastercard2024} capabilities. 
However, this system operates exclusively within the Mastercard payment ecosystem and does not provide a generalized framework.


In addition, our recent case study~\cite{case2025kim} reveals that current e-commerce websites lack a standardized model to delegate transactional authority to agentic AI.
Specifically, most existing e-commerce platforms do not meet the operational needs of agentic AI, such as setting user-defined constraints, performing automated purchases, and handling logins without human intervention. 
This clearly demonstrates the need for agent-aware website design principles.

To address these gaps, this paper proposes an open-source software approach toward fine-grained access-controlled (authorized) website interaction for agentic AI carrying out critical tasks securely delegated by human users.
Our technical contributions include:
\begin{compactItemize}
\item Extension of an open-source authorization toolkit for distributed systems to enable secure access delegation to AI agents, by secure delivery of cryptographic tokens for access-controlled interaction between AI agents and websites.
\item Design of website interface for human users to specify fine-grained access control for access delegation, as well as for AI agents to perform critical tasks delegated by human users.
\item Open-source release of full design of our proposed extension and website, followed by evaluation and validation of our design, with in-depth discussions for future directions.
All implementation artifacts of this paper are publicly available on GitHub\footnote{\hyperlink{https://github.com/asu-kim/agentic-website}{https://github.com/asu-kim/agentic-website}, these artifacts include source code for the frontend and backend of our website, agent-side scripts, and experimental logs.}.


\end{compactItemize}





%

\section{Proposed Approach}

\begin{figure}
\centering
\includegraphics[width=0.9\linewidth]{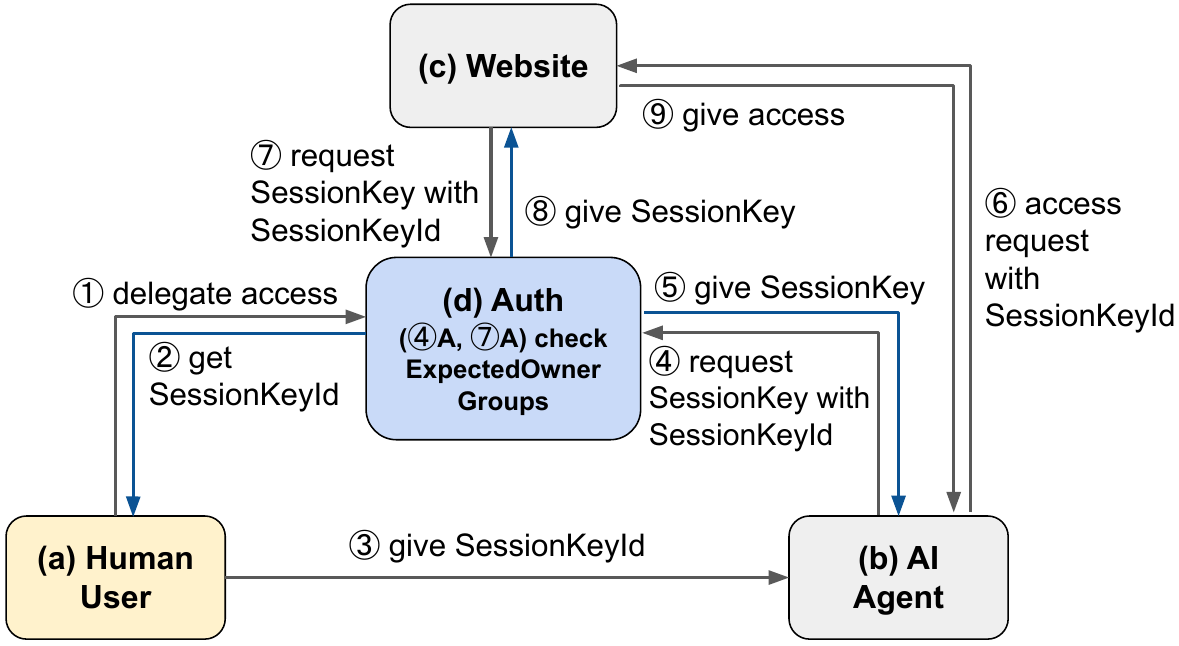}
\vspace{-12pt}
\caption{Our proposed system's workflow for secure access delegation of our approach, where the user grants delegated access to the agent for accessing the website.}
\Description{The user delegates access to an agent, which retrieves a session key from Auth. The website independently verifies the session key with Auth.}
\vspace{-10pt}
\label{fig:overview}
\end{figure}

\figurename~\ref{fig:overview} shows an overview of our approach to enable secure access delegation among (a) human users, (b) AI agents and (c) a website using (d) \emph{Auth}~\cite{kim2016secure}, a key entity of an open-source toolkit for decentralized authorization services, \emph{Secure Swarm Toolkit (SST)}~\cite{kim2017toolkit}.
An Auth acts as a key distribution center (KDC)~\cite{neuman1994kerberos,kim2017authentication} for issuing cryptographic session keys for secure, authorized (access-controlled) communication with specified validity periods. 


\begin{figure}
  \centering
  \includegraphics[width=1.0\linewidth]{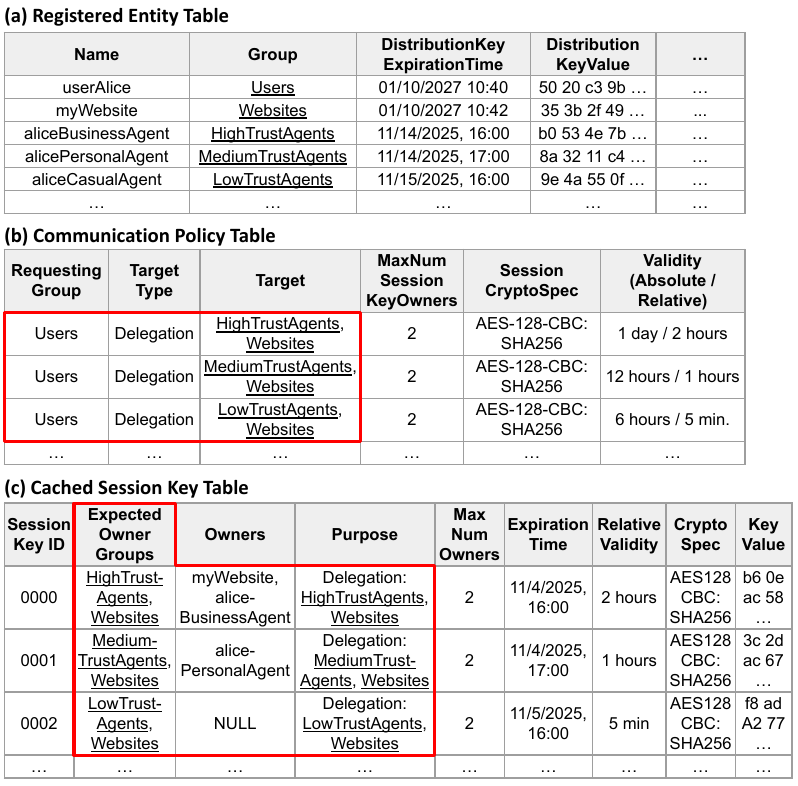}
\vspace{-22pt}
  \caption{Extended Auth database tables supporting secure access delegation with example data rows.}
  \Description{We added a new target type, Delegation, in table (b), defining which agent groups and the website can receive delegated access from a user. We also added the ExpectedOwnerGroups in table (c), which restricts which entities are allowed to obtain a session key.}
  \label{fig:table1}
    \vspace{-10pt}
\end{figure}

\subsection{Secure Access Delegation}

An Auth~\cite{kim2016secure} uses local database tables to store credentials and the necessary information to authenticate and authorize network entities.
Since the existing design of Auth and SST~\cite{kim2017toolkit} does not support access delegation, in this paper, we extend Auth's database schema and data types to enable secure access delegation.

\figurename~\ref{fig:table1} describes the Auth's key database tables and our extension (highlighted in red boxes) using user and AI agent examples.
\begin{compactItemize}
\item \textbf{Registered entity table (\figurename~\ref{fig:table1}a)} stores credentials and unique string identifiers for each network entity, the group the entity belongs to, as well as cryptography and validity periods of distribution keys, which are used for secure delivery of \emph{cached session keys} to be used for access control of entities.

\begin{compactItemizeEmptyBullet}
\item Our example shown in \figurename~\ref{fig:table1}a includes a user named Alice (\texttt{userAlice}), a website (\texttt{myWebsite}), and Alice's three AI agents (\texttt{Business}, \texttt{Personal}, \texttt{Casual}), belonging to groups, \texttt{Users}, \texttt{Websites}, and \texttt{High/Medium/LowTrustAgents}.
\end{compactItemizeEmptyBullet}

\item \textbf{Communication policy table (\figurename~\ref{fig:table1}b)} governs the access between entities, for example, which entity can gain access to which entity (or which group of entities), as well as the type of the communication target, crypto, and access duration (relative validity starting from the first access and absolute validity as a hard deadline for the last access attempt). Auth always consults with this table before granting access in the form of session keys.

\begin{compactItemizeEmptyBullet}
\item \textbf{Our extension in this paper} introduces a new target type, \texttt{Delegation}, to specify that the requesting group that initiates access delegation, while not being part of the communication.
In \figurename~\ref{fig:table1}b, for example, the requesting group is \texttt{Users}, which Alice belongs to.
Also, our new target type describes two entity groups using delegated access, with the first group (e.g., agents with \texttt{High/Medium/Low} trust) granted delegated access by \texttt{Users} to access the second group (e.g., \texttt{Websites}).
For example, Alice can request Auth for delegated access for one of her agents to a target website.
\end{compactItemizeEmptyBullet}

\item \textbf{Cached session key table (\figurename~\ref{fig:table1}c)} stores cryptographic keys as tokens for granted accesses among network entities.
This table stores each session key's information, including the current owners, unique ID, designated purpose, and expiration time.
This table refers to a session key as ``\emph{cached}'' because Auth temporarily stores these keys for granted accesses so that Auth can provide the session key for other requesting entities.
\begin{compactItemizeEmptyBullet}
\item \textbf{Our extension} adds a new column, \texttt{ExpectedOwnerGroups}, as shown in \figurename~\ref{fig:table1}c, which is populated when a session key for delegated access is created, referring to the \texttt{Target} column of the communication policy table.
Auth ensures that each owner is from each group as specified in the expected owner groups for secure access delegation. 
\end{compactItemizeEmptyBullet}
\end{compactItemize}



\noindent

\noindent
We walk through the access delegation process of our proposed system in \figurename~\ref{fig:overview} using the example database tables in \figurename~\ref{fig:table1}.

A user initiates delegation (\figurename~\ref{fig:overview}\textcircled{1}) by requesting Auth to create a session key for an agent, providing the agent’s trust level.
Auth checks the communication policy table for the appropriate validity period for the given trust level and generates a new session key to be stored in the cached session key table with an initially empty \texttt{Owner} and returns the session key's ID to the user (\figurename~\ref{fig:overview}\textcircled{2}).

Upon receiving the session key ID from the user (\figurename~\ref{fig:overview}\textcircled{3}), the agent requests the corresponding session key from Auth (\figurename~\ref{fig:overview}\textcircled{4}), Auth checks whether the requesting agent belongs to one of the session key’s \texttt{ExpectedOwnerGroups} (\figurename~\ref{fig:overview}\textcircled{4}A).
If the agent is authorized but not yet an owner, Auth returns the session key and registers the agent as the owner (\figurename~\ref{fig:overview}\textcircled{5}). 
If the agent is not in the \texttt{ExpectedOwnerGroups}, Auth denies the request. 
Similarly, if an already-registered agent attempts to re-request the key, Auth rejects it to prevent duplicate issuance. 
Through this, Auth ensures that delegated session keys are issued exactly once to each authorized agent and never shared with unauthorized ones.

\subsection{Website Design}


We design the website to provide secure and fine-grained access delegation for autonomous agents acting on behalf of a human user.
We implement the frontend in JavaScript using a React\footnote{\hyperlink{https://react.dev/}{https://react.dev/}} framework, and build the backend with Python Flask\footnote{\hyperlink{https://flask.palletsprojects.com/en/stable/}{https://flask.palletsprojects.com/en/stable/}}.

\begin{figure}
  \centering
  \includegraphics[width=\linewidth]{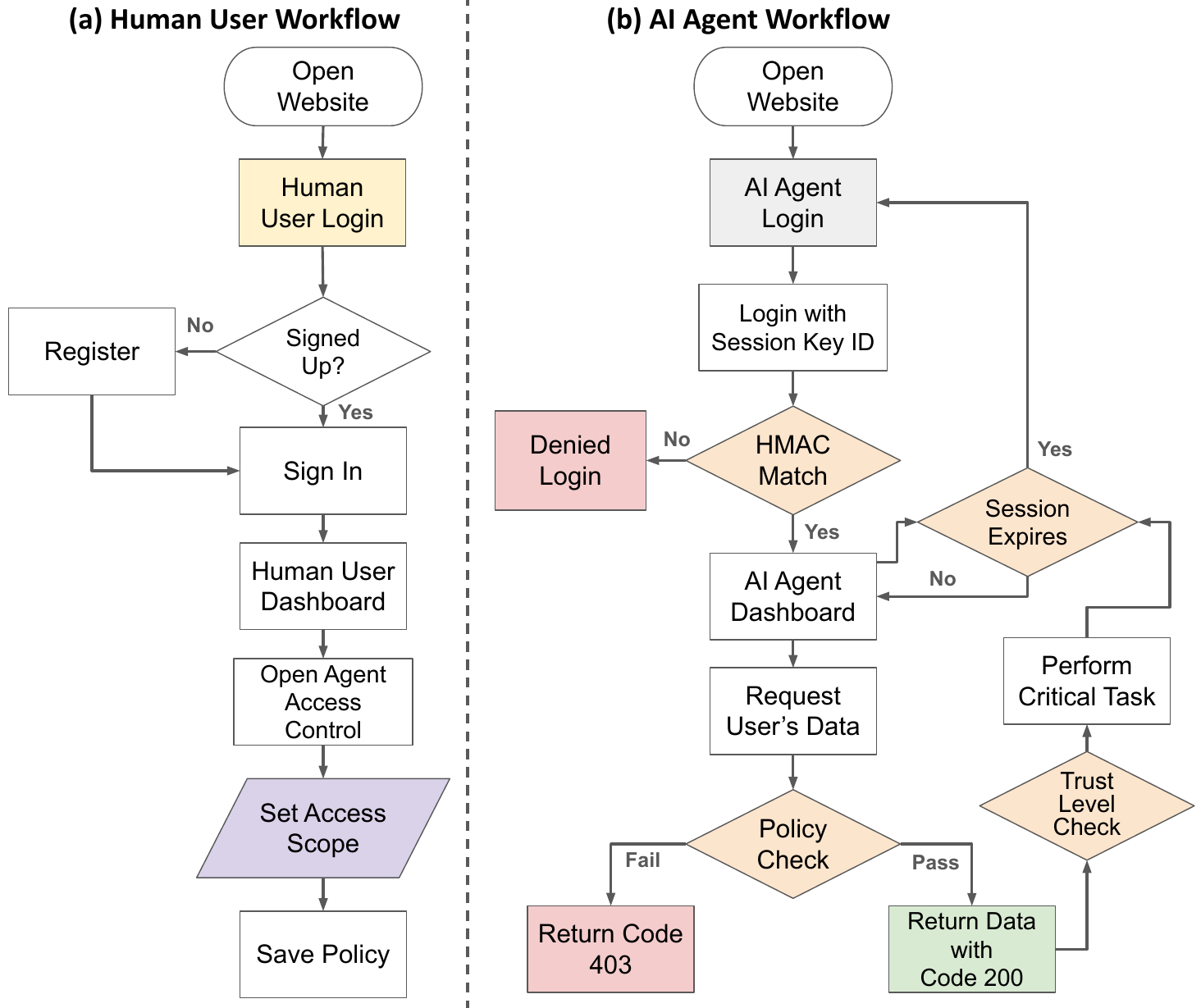}
  \vspace{-20pt}
  \caption{Website workflows from the perspective of (a) human users and (b) AI agents.}
  \Description{Description}
  \label{fig:website_overview}
  \vspace{-12pt}
\end{figure}

\figurename~\ref{fig:website_overview} shows the workflow of our website.
Human users configure the access scopes of their agents, and each agent interacts with the website through a verifiable authentication workflow coordinated with the Auth entity.
The website's role is to enforce these user-defined policies, ensuring that only authorized agents can operate within their assigned boundaries.

\begin{figure}
  \centering
  \vspace{-12pt}
  \includegraphics[width=0.7\linewidth]{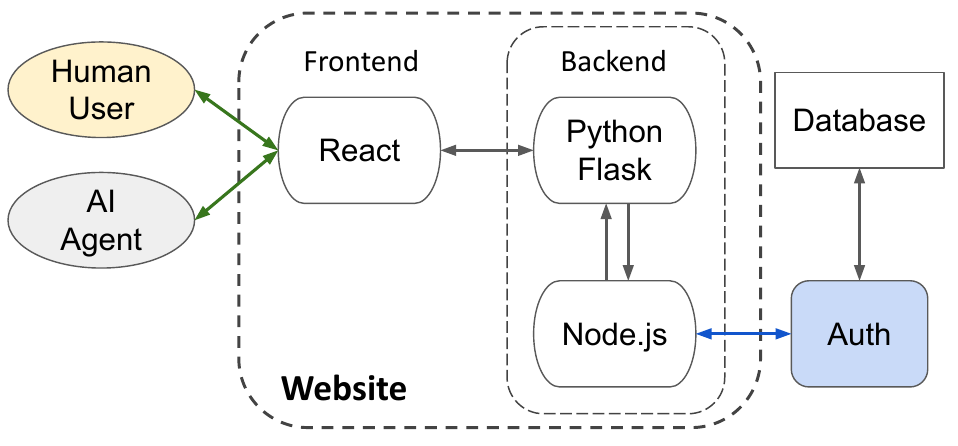}
  \vspace{-12pt}
  \caption{Overview of our website design.}
  \Description{Description}
  \label{fig:overview_structure}
  \vspace{-10pt}
\end{figure}

When an agent attempts to log in, the website generates and displays a 32-digit random nonce.
The agent then computes a keyed-hash message authentication code (HMAC)~\cite{fips198} using the website-generated nonce and the session key obtained at (\figurename~\ref{fig:overview}\textcircled{5}), and submits both the session key ID and the resulting HMAC value as its proof of authentication (\figurename~\ref{fig:overview}\textcircled{6}).

Upon receiving these values, the website contacts Auth to obtain the corresponding session key (\figurename~\ref{fig:overview}\textcircled{7}). 
Auth verifies the session key ID and the requesting entities authorization (\figurename~\ref{fig:overview}\textcircled{7}A) and returns both the session key and the agent’s group information, which was registered as the key’s owner during the delegation process in Section~2.1. 
After receiving the session key (\figurename~\ref{fig:overview}\textcircled{8}), the website calculates the HMAC value over the same nonce. 
Authentication succeeds only if the HMAC values generated by the agent and the website match, confirming that the agent genuinely possesses the valid session key issued by Auth (\figurename~\ref{fig:overview}\textcircled{9}).

After successful authentication, the website uses the agent's group name and the relative validity period provided by Auth to determine which data the agent can access and to configure the session lifetime.
The relative validity period specifies how long the agent may operate before re-authentication is required.

When the agent issues a data request, the website evaluates the request against the user-configured access-control policy. 
If the agent attempts to access information outside its authorized scope, the website rejects the request and returns an error.


With the retrieved user's data, the agent can perform a mock critical task on the website. 
For demonstration purposes, we include a simulated purchase workflow on the website as the critical operation that the agent can execute.
In the critical task interface, the agent enters the item name provided by the user and uses the user’s address, card information, and phone number retrieved from the website. 
The agent can then perform the mock purchase, showing how a highly trusted agent handles sensitive delegated operations.

By combining user-defined policy enforcement, HMAC verification with the session key, and trust-based session management, the website ensures that autonomous agents operate strictly within their delegated authority.

\begin{table}[]
\caption{Summary of aspects validated by our evaluation.}
  \vspace{-7pt}
\centering
\footnotesize
\label{tab:evaluation}
\begin{tabular}{|l|l|l|l|}
\hline
\textbf{\begin{tabular}[c]{@{}l@{}}Evaluation \\ Aspect\end{tabular}}               & \textbf{Test Scenario}                                                                                 & \textbf{\begin{tabular}[c]{@{}l@{}}Expected \\ Behavior\end{tabular}}                          & \textbf{Observed Results}                                                                                \\ \hline
\textbf{Authentication}                                                             & \begin{tabular}[c]{@{}l@{}}Agent computes \\ HMAC with \\ correct session key\end{tabular}             & \begin{tabular}[c]{@{}l@{}}Login succeeds \\ only with \\ valid key\end{tabular}               & \begin{tabular}[c]{@{}l@{}}100\% success \\ with correct key; \\ 0\% with invalid key\end{tabular}       \\ \hline
\textbf{\begin{tabular}[c]{@{}l@{}}Fine-Grained \\ Access Control\end{tabular}}     & \begin{tabular}[c]{@{}l@{}}Low-trust agent \\ requests email and \\ phone fields\end{tabular}          & \begin{tabular}[c]{@{}l@{}}Allowed: email; \\ Denied: phone\end{tabular}                       & \begin{tabular}[c]{@{}l@{}}email returned 200; \\ phone blocked 403\end{tabular}                         \\ \hline
\textbf{\begin{tabular}[c]{@{}l@{}}Unauthorized \\ Access \\ Handling\end{tabular}} & \begin{tabular}[c]{@{}l@{}}Invalid \\ agent requests \\ session key or \\ repeats request\end{tabular} & \begin{tabular}[c]{@{}l@{}}Auth rejects \\ unauthorized \\ or repeated \\ request\end{tabular} & \begin{tabular}[c]{@{}l@{}}Unauthorized: \\ 100\% refused; \\ Repeated request: \\ 100\% rejected\end{tabular} \\ \hline
\textbf{\begin{tabular}[c]{@{}l@{}}Session \\ Management\end{tabular}}              & \begin{tabular}[c]{@{}l@{}}Session used \\ after relative \\ validity period\end{tabular}              & \begin{tabular}[c]{@{}l@{}}Session should \\ be terminated \\ automatically\end{tabular}       & \begin{tabular}[c]{@{}l@{}}100\% \\ Auto terminated\end{tabular}                                         \\ \hline
\end{tabular}
  \vspace{-12pt}
\end{table}

\vspace{-5pt}
\section{Evaluation}
As shown in \tablename~\ref{tab:evaluation}, to evaluate our approach, we focus on validating four key aspects of the system: authentication correctness, fine-grained access control, handling unauthorized access, and session management.
The experiments are implemented in Python 3 using OpenAI’s gpt-oss-20b model~\cite{openaigpt2025} for agent-side reasoning and decision-making.
As shown in \figurename~\ref{fig:website_overview} (a), the system, including the agent, website, and SST's Auth, is deployed on the ASU Sol Supercomputer~\cite{asu} equipped with an NVIDIA A100 GPU, while the website interface is accessed through the Firefox browser.

\vspace{-3pt}
\paragraph{\textbf{Authentication via Login Process}}
We verify that authentication succeeds only when the agent has the correct session key value. 
Using the session key retrieved from Auth, the agent computes the HMAC over the website-generated nonce. 
Login succeeds consistently when the session key is valid. 

\vspace{-3pt}
\paragraph{\textbf{Fine Grained Access Control}}
We test fine-grained access control using agents with different trust levels. 
Agents with permissions limited to specific fields (e.g., email) successfully retrieve only the allowed data, while all attempts to access higher-privilege fields (e.g., phone or address) are rejected. 
These results confirm that user-configured, per-agent scopes are enforced correctly.

\vspace{-3pt}
\paragraph{\textbf{Unauthorized Access Handling}}
We evaluate how the system handles incorrect or unauthorized requests issued toward Auth. 
If an agent requests a session key twice using the same session key Id, Auth correctly recognizes the repeated request and denies it. 
More importantly, when an agent outside the \texttt{ExpectedOwnerGroups} attempts to obtain a session key, Auth consistently rejects the request. 
This ensures that agents cannot impersonate one another or escalate privileges through unauthorized key requests.

\vspace{-3pt}
\paragraph{\textbf{Session Management}}
To validate temporal security policies, we set the session expiration time using the relative validity period embedded in the session key. 
After the duration expires, the website automatically invalidates the session and logs out the agent. 
This behavior confirms that time-based trust enforcement works predictably and prevents sessions from being reused.

\vspace{-3pt}
\paragraph{\textbf{Summary}}
We measure the success rate of the four evaluation aspects across five trials for each condition and agent trust level.
All valid agents consistently achieve a 100\% success rate in authentication, authorized access, and correct session termination after expiration.
In contrast, invalid operations such as using incorrect session keys, attempting to obtain session keys outside the \texttt{ExpectedOwnerGroups}, or issuing repeated session key requests consistently fail with a 0\% success rate.

\vspace{-3pt}
\paragraph{\textbf{Latency Measurement}}

\begin{figure}[t]
\centering
    \begin{subfigure}[b]{0.43\linewidth}
    \centering
    \footnotesize
    \begin{tabular}{|l|c|c|}
    \hline
    \textbf{Case} & \textbf{\begin{tabular}[c]{@{}l@{}}Average\\(ms)\end{tabular}} & \textbf{\begin{tabular}[c]{@{}l@{}}Std.\\(ms)\end{tabular}} \\ \hline
    \textbf{\begin{tabular}[c]{@{}l@{}}High-trust \\ agent (success)\end{tabular}}& 127{,}248.8 & 11{,}330.0 \\ \hline
    \textbf{\begin{tabular}[c]{@{}l@{}}Medium-trust \\ agent (success)\end{tabular}}& 140{,}060.2 & 17{,}407.0 \\ \hline
    \textbf{\begin{tabular}[c]{@{}l@{}}Low-trust \\ agent (success)\end{tabular}}& 137{,}842.4 & 6{,}733.4 \\ \hline
    \textbf{\begin{tabular}[c]{@{}l@{}}Unauthorized\end{tabular}}& 98{,}690.7  & 11{,}780.6 \\ \hline
    \end{tabular}
    \vspace{-5pt}
    \caption{Latencies with average and standard deviation (std.).}
    \label{fig:latency_table}
    \end{subfigure}
    \hfill
    \begin{subfigure}[b]{0.45\linewidth}
    \centering
    \includegraphics[width=\linewidth]{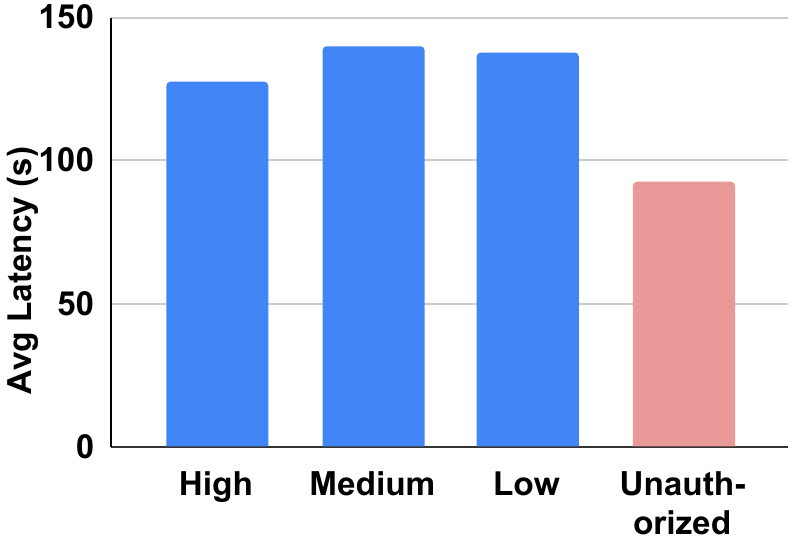}
    \vspace{-5pt}
    \caption{Latency comparison chart.}
    \label{fig:latency_chart}
    \end{subfigure}
    \vspace{-10pt}
\caption{Comparison of delegated-access latency across trust levels and unauthorized requests.}
\label{fig:latency_combined}
\vspace{-17pt}
\end{figure}

Additionally, we measure the end-to-end latency of the delegated access workflow, defined as the time from the beginning of the agent’s execution to the moment the agent successfully retrieves the permitted data from the website.

As summarized in \figurename~\ref{fig:latency_combined}, the end-to-end average latency of unauthorized key requests (98.7 seconds) is significantly lower than the end-to-end average latency of successful delegated access (127–140 seconds). This is because Auth rejects session key requests from entities not in the \texttt{ExpectedOwnerGroups}, and the agent cannot proceed with any website interaction.
The medium- and low-trust agents show slightly higher latency due to variability in the agent-side LLM inference time, not due to the trust levels.

Since our system is deployed on the local host on the ASU Sol supercomputer~\cite{asu}, the measured end-to-end latency does not include network latency. 
In a distributed deployment, the total end-to-end latency would additionally depend on round-trip communication between the agent, Auth, and the website.
Assuming symmetrical network latencies, the total latency can be represented as,
\vspace{-3pt}
\begin{equation}
L_{total} = L_{e2e} + 4\times{}L_{a2A} + 4\times{}L_{w2A} + x\times{}L_{a2w}
\label{eq:latency}
\vspace{-3pt}
\end{equation}

where $L_{e2e}$ is the end-to-end latency measured in \figurename~\ref{fig:latency_combined}, $L_{a2A}$ and $L_{w2A}$ are latencies between the agent/website and Auth, respectively, and $L_{a2w}$ is the agent-website latency with $x$ for the number of agent-website interactions.
This $x$ is expressed as $x = 2 + n$, where $2$ is for one request to load the login page and another request to submit the HMAC value for authorization, whereas $n$ is for $n$ additional requests corresponding to retrieving the number of sensitive data items.

\vspace{-3pt}
\section{Discussion}

By leveraging SST~\cite{kim2017toolkit}, an open-source and fully decentralized key distribution center (KDC), our design avoids reliance on third-party cloud infrastructure, allowing the proposed system to operate entirely locally and independently.
This design enables fine-grained, per-agent permissions, managed by individual entities, such as households, companies, schools, or hospitals.
Our approach can be generalized beyond AI agents and support any delegated-authority configuration involving human-to-entity or entity-to-entity delegation, tailored to an autonomous delegation for various scenarios.

Although the proposed approach has been implemented and functionally tested, formal verification remains future work.
Properties such as correct ownership enforcement, enforcement of access scope boundaries, and consistency between website-side and Auth-side session-key checks can be modeled and validated using formal verification tools, such as Alloy\footnote{\url{https://alloytools.org/}}~\cite{jackson2012software}, to guarantee no authorization violations occur under any execution or communication path.


\begin{acks}
This work was supported in part by the NSF grant POSE-\#2449200.
\end{acks}

\vspace{-2pt}
\bibliographystyle{ACM-Reference-Format}
\bibliography{software}

@article{khan2024security,
  title={Security threats in agentic {AI} system},
  author={Khan, Raihan and Sarkar, Sayak and Mahata, Sainik Kumar and Jose, Edwin},
  journal={arXiv preprint arXiv:2410.14728},
  year={2024}
}

@misc{pymnts2024,
  author = {PYMNTS},
  title = {{AI} To Power Personalized Shopping Experiences In 2025},
  year = {2024},
  url = {https://www.pymnts.com/artificial-intelligence-2/2024/ai-to-power-personalized-shopping-experiences-in-2025/},
  note = {{Accessed}: 2025-08-14}
}

@article{south2025authenticated,
  title={Authenticated delegation and authorized {AI} agents},
  author={South, Tobin and others},
  journal={arXiv preprint arXiv:2501.09674},
  year={2025}
}

@article{huang2025novel,
  title={A novel zero-trust identity framework for agentic {AI}: Decentralized authentication and fine-grained access control},
  author={Huang, Ken and others},
  journal={arXiv preprint arXiv:2505.19301},
  year={2025}
}

@misc{mastercard2025,
  author       = {{Mastercard}},
  title        = {Mastercard unveils {Agent Pay}, pioneering agentic payments technology to power commerce in the age of {AI}},
  url = {https://www.mastercard.com/us/en/news-and-trends/press/2025/april/mastercard-unveils-agent-pay-pioneering-agentic-payments-technology-to-power-commerce-in-the-age-of-ai.html},
  year         = {2025},
  note         = {Accessed: 2025-08-20}
}

@misc{mastercard2024,
  author = {{Mastercard}},
  title = {Tokenization explained: Protecting sensitive data and strengthening every transaction},
  url = {https://www.mastercard.com/us/en/news-and-trends/stories/2025/what-is-tokenization.html},
  year = {2024},
  note = {Accessed: 2025-08-20}
}

@misc{openai2025,
  author = {{OpenAI}},
  title = {Introducing {AgentKit}},
  url = {https://openai.com/index/introducing-agentkit/},
  year = {2025},
  note = {Accessed: 2025-10-10}
}

@inproceedings{case2025kim,
    author = {Kim, Sunyoung and Kim, Hokeun},
    title = {A Case Study on Delegating Critical Tasks to Agentic {AI} and Prototype Access Control Methods},
    booktitle = {the IEEE 5th Cyber Awareness and Research Symposium 2025 (CARS'25)},
    year = 2025
}

@inproceedings{kim2017toolkit,
  title={A toolkit for construction of authorization service infrastructure for the {Internet of Things}},
  author={Kim, Hokeun and Kang, Eunsuk and Lee, Edward A and Broman, David},
  booktitle={Proceedings of the Second International Conference on Internet-of-Things Design and Implementation},
  pages={147--158},
  year={2017}
}

@misc{openaigpt2025,
  author = {{OpenAI}},
  title = {Introducing gpt-oss},
  url = {https://openai.com/index/introducing-gpt-oss/},
  year = {2025},
  note = {Accessed: 2025-11-14}
}

@inproceedings{kim2016secure,
  title={A secure network architecture for the internet of things based on local authorization entities},
  author={Kim, Hokeun and Wasicek, Armin and Mehne, Benjamin and Lee, Edward A},
  booktitle={2016 IEEE 4th International Conference on Future Internet of Things and Cloud (FiCloud)},
  pages={114--122},
  year={2016},
  organization={IEEE}
}

@misc{asu,
  author = {{Arizona State University Core Research Facilities}},
  title = {Computing and Data Service},
  url = {https://cores.research.asu.edu/research-computing/},
  year = {Accessed on 2025-11-14}
}

@book{jackson2012software,
  title={Software Abstractions: logic, language, and analysis},
  author={Jackson, Daniel},
  year={2012},
  publisher={MIT press}
}

@inproceedings{zhang2024multimodal,
  title={A multimodal foundation agent for financial trading: Tool-augmented, diversified, and generalist},
  author={Zhang, Wentao and others},
  booktitle={Proceedings of the 30th ACM SIGKDD Conference on Knowledge Discovery and Data Mining},
  pages={4314--4325},
  year={2024}
}

@article{moritz2025coordinated,
  title={Coordinated {AI} agents for advancing healthcare},
  author={Moritz, Michael and Topol, Eric and Rajpurkar, Pranav},
  journal={Nature Biomedical Engineering},
  pages={1--7},
  year={2025},
  publisher={Nature Publishing Group UK London}
}

@article{neuman1994kerberos,
  title={Kerberos: An authentication service for computer networks},
  author={Neuman, B Clifford and Ts'o, Theodore},
  journal={IEEE Communications magazine},
  volume={32},
  number={9},
  pages={33--38},
  year={1994},
  publisher={IEEE}
}

@article{kim2017authentication,
  title={Authentication and Authorization for the {Internet of Things}},
  author={Kim, Hokeun and Lee, Edward A},
  journal={IT Professional},
  volume={19},
  number={5},
  pages={27--33},
  year={2017},
  publisher={IEEE}
}

@article{fips198,
year={2008},
  title={Federal information processing standards publication: The keyed-hash message authentication code (HMAC)},
  author={{FIPS PUB 198-1}},
  journal={Information Technology Labratory, National Institute of Standards and Technology (NIST), Gaithersburg, MD 20899--8900}
}


\end{document}